\def\year{2022}\relax
\documentclass[letterpaper]{article} 
\usepackage{aaai22}  
\usepackage{times}  
\usepackage{helvet}  
\usepackage{courier}  
\usepackage[hyphens]{url}  
\usepackage{graphicx} 
\usepackage{booktabs}
\usepackage{subfig}
\usepackage{multirow}
\usepackage{amsmath}
\usepackage{natbib}  
\usepackage{caption} 
\DeclareCaptionStyle{ruled}{labelfont=normalfont,labelsep=colon,strut=off} 
\usepackage{algorithm}
\usepackage{algorithmic}

\usepackage{newfloat}
\usepackage{listings}
\floatstyle{ruled}
\newfloat{listing}{tb}{lst}{}
\floatname{listing}{Listing}

\DeclareMathOperator*{\argmax}{arg\,max}

\makeatletter
\newcommand{\printfnsymbol}[1]{%
  \textsuperscript{\@fnsymbol{#1}}%
}
\makeatother

\usepackage{url}

\title{Grounding Natural Language Instructions: Can Large Language Models Capture Spatial Information?}

\author{
Julia Rozanova\thanks{Equal contribution.}$^1$, Deborah Ferreira\printfnsymbol{1}$^1$, Krishna Dubba$^3$,\\ Weiwei Cheng$^3$, Dell Zhang$^4$, Andre Freitas$^{12}$\\
 $^1$Department of Computer Science, University of Manchester, UK \\
 $^2$Idiap Research Institute, Switzerland \\
 $^3$Blue Prism AI Labs, London, UK\\
 $^4$ByteDance AI Lab, London, UK \\
 \{julia.rozanova, deborah.ferreira,  andre.freitas\}@manchester.ac.uk, \{krishna.dubba, weiwei.cheng\}@blueprism.com, dell.z@ieee.org}

\date{}

\begin{document}
\maketitle
\begin{abstract}

Models designed for intelligent process automation are required to be capable of grounding user interface elements. 
This task of \emph{interface element grounding} is centred on linking instructions in natural language to their target referents. 
Even though BERT and similar pre-trained language models have excelled in several NLP tasks, their use has not been widely explored for the UI grounding domain.
This work concentrates on testing and probing the grounding abilities of three different transformer-based models: BERT, RoBERTa and LayoutLM.
Our primary focus is on these models' spatial reasoning skills, given their importance in this domain. We observe that LayoutLM  has a promising advantage for applications in this domain, even though it was created for a different original purpose (representing scanned documents): 
the learned spatial features appear to be transferable to the UI grounding setting, 
especially as they demonstrate the ability to discriminate between target \emph{directions} in natural language instructions.

\end{abstract}
\section{Introduction}

The majority of human-computer interactions are mediated by Graphical User Interfaces (GUIs) within desktop and mobile environments. 
GUIs provide the interaction backbone in which end-users compose and repurpose atomic functional elements embedded within existing applications. 

Recently, there is an emerging effort to build models to support the automation of long-tail processes within a multi-application Desktop/GUI environment~\cite{ferreira2020evaluation}.
This effort encompasses the interfacing of Human-Desktop Interaction (HDI) with natural language, both for HDI understanding (the creation of natural language descriptions that describe HDIs) and HDI semantic parsing (the execution of HDI actions from commands expressed in natural language). 

One fundamental skill required for designing such models is being able to \emph{ground} UI element mentions in natural language instructions, in the style of a mapping task.
A natural language command in most cases will contain an action verb.
Often, this verb has as an argument an object that refers to an interface element.
A model geared towards understanding this instruction should be aware of the mapping between words and interface elements' concrete realisation.
This mapping is named \textit{interface element grounding task}~\cite{pasupat2018mapping} or UI grounding.
Consider the instruction ``Click on the cancel button.'': in order to ground this instruction, one is expected to find the cancel button in a screen. Applications of a UI grounding model ranges from intelligent process automation to the creation of accessibility tools.

Transformer-based models have achieved excellent performance in tasks requiring semantic inference, with state of the art results in various NLP tasks. However, such models' expertise in reasoning with elements positioned in a 2-dimensional space has limited research, especially in the interface element grounding domain.

This work focuses on testing and probing the grounding abilities of three different transformer-based models, namely BERT~\cite{devlin-etal-2019-bert}, RoBERTa~\cite{liu2019roberta}, and LayoutLM \cite{xu2020layoutlm}.
This process is performed in two parts: first, evaluating the accuracy for UI grounding for different reasoning types, and second, probing the generated representations using spatial auxiliary tasks.

The contributions of this work can be summarised as follows:

\begin{itemize}
\item We provide a first systematic study of the spatial reasoning capabilities of transformer-based models in the UI grounding setting.
\item We compare different pre-trained transformer-based architectures and present a probing-based analysis of their spatial reasoning capabilities.
\item We empirically demonstrate that the semantic representation of textual and layout information from scanned documents can also be successfully transferred to represent user interfaces.
\item We propose a novel application domain for transformer-based models.
\item We demonstrate the need for more complex spatial reasoning datasets and models capable of relative spatial reasoning.

\end{itemize}

\section{Problem Formulation: Grounding Interface Elements}

Given a command (noun phrase) $c$ contained in a natural language command and an user interface composed of a set $U = \{ u_1, u_2, \ldots, u_2\}$ of \emph{interface elements}, where each element $i$ is composed of a given text content $u_{i_T}$ and a bounding box $u_{i_{BB}}=[u_{i_{x_0}},u_{i_{x_1}},u_{i_{y_0}}, u_{i_{y_1}}]$, the \textit{grounding task} aims to select the correct element $u_i \in U$ which is referred in phrase $c$, assuming that this phrase only refers to a unique element.

\noindent Given $U$ and $c$, the following condition distribution over the elements is defined:

\begin{equation}
    p(u_i|p) \propto exp[s(f(c), g(u_i))]
\end{equation}

\noindent where $g(\cdot)$ and $f(\cdot)$ are embedding functions for the UI elements and phrase, respectively, and $s(\cdot, \cdot)$ is a scoring function between command and UI element. To obtain the correct interface element, one needs to maximise the log-likelihood for the correct interface element.

To solve the grounding task, a model needs mainly to focus on three types of reasoning~\cite{li2020mapping, pasupat2018mapping}, subject to what is required by the command $c$. The types of reasoning are \emph{extractive}, \emph{absolute spatial} and \emph{relative spatial}.  

\begin{figure}[h]
    \centering
    \includegraphics[width=0.85\linewidth]{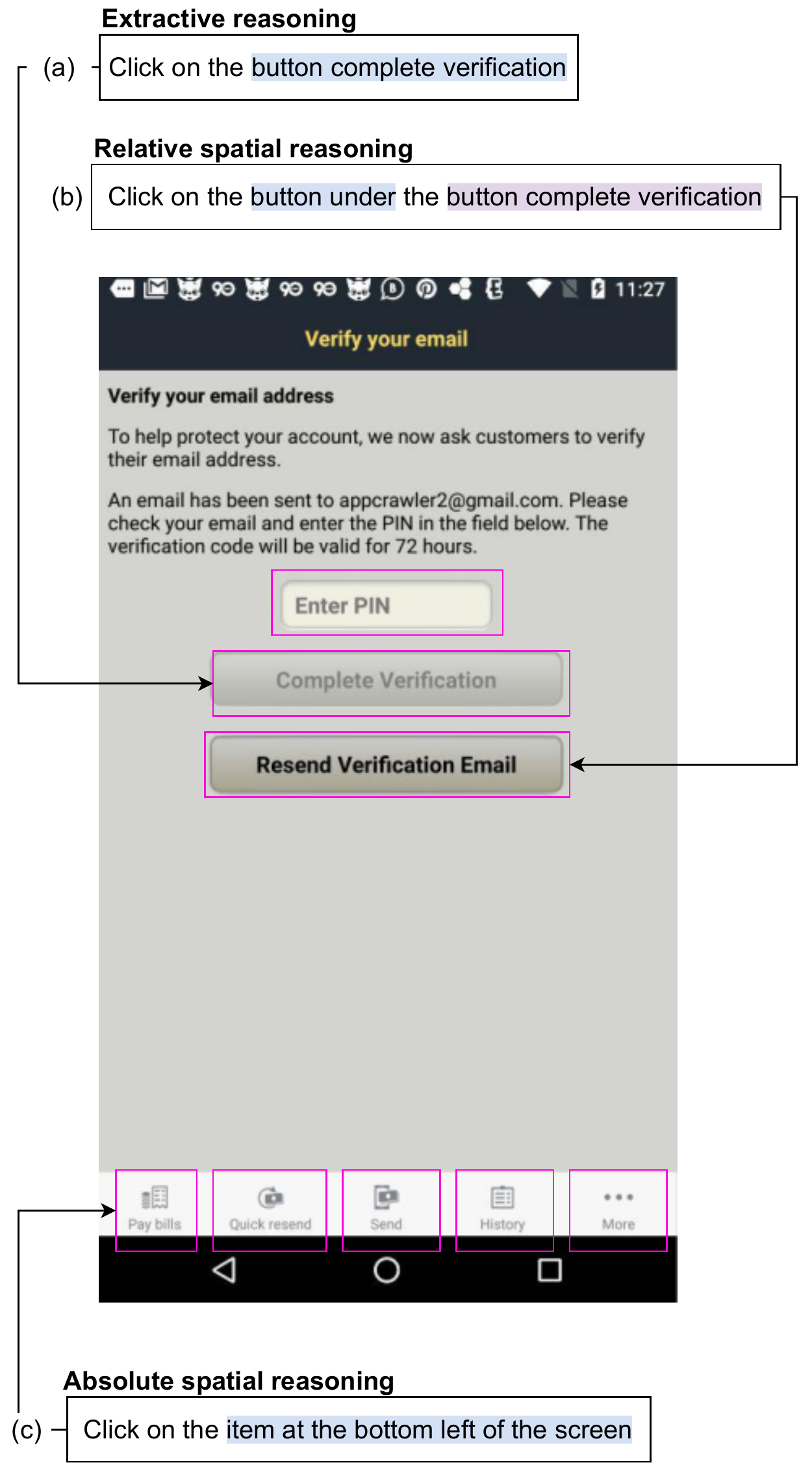}
    \caption{Example of three different commands, requiring different types of reasoning, and matching interface elements in a mobile screen. The screen presented here is part of the Rico dataset~\cite{Deka:2017:Rico}.}
    \label{fig:example_rico}
\end{figure}

Extractive reasoning is required in the most trivial cases, where one needs only to match the textual content $u_{i_T}$ of the interface element with the text in $c$. This scenario includes cases where there is an exact keyword matching and also where some paraphrasing is required. As shown in example (a) from Figure~\ref{fig:example_rico}, for this type of reasoning, a model does not require any spatial information about the interface element. 

When spatial reasoning is needed, the task increases in complexity. Absolute spatial reasoning requires a model to understand the positional information contained in the command and to be able to locate the interface element inside a screen. Figure~\ref{fig:example_rico}(b) presents an example of this case, where it is not possible to identify the element without having the values of its bounding box. However, it does not require the representation of surrounding interface elements, unlike relative spatial reasoning.

Relative spatial reasoning is the most challenging type considering a model is expected to interpret a relative spatial command, demanding an understanding of the screen as a whole. In this type of command, an interface element is referred to in relation to another element. As shown in Figure~\ref{fig:example_rico}(c), this type of instruction can be easily mistaken with an extractive reasoning command in circumstances where a model is not knowledgeable of the spatial nature of the task given then extra noise with textual information from a different element.

\section{Related Work}

\subsection{UI Grounding}
Past research on grounding interface elements has mainly focused on applications to mobile screens and web elements.
\citet{li2020mapping} proposes Seq2Act, a two-stage Transformer model for grounding UI elements in an Android environment.
In our work, we compare pre-trained models with the Seq2Act model.
\citet{pasupat2018mapping} proposes an embedding based model for grounding instructions based on web elements.
They are able to represent not only positional information but also properties of UI elements. 

There is still limited research in the task of interface element grounding. However, a considerable amount of literature has been published on semantic parsing for task execution.
\citet{goldwasser2011learning} and \citet{volkova2013lightly} focus on using natural language instructions to learn the concepts and rules of an environment, where these elements can be used to execute and verify tasks.
Similarly, the work by \citet{branavan2012learning} use text from manuals to learn the rules of the game Civilization II.

For goal-oriented tasks, the parsing can be modelled around a reward/punishment function.
\citet{branavan2010reading} explicitly models an environment where a reward function is used to evaluate how well a command sequence achieves a described task.
\citet{macglashan2014training} propose a learning algorithm for generating high-level task definitions from instruction, derived from online human-delivered reward and punishment.

\subsection{Probing}
There is often interest in the \emph{intermediate} features or information captured by neural models. 
There are two broad categories of approaches to determining the extent to which certain features or
phenomena are captured: either \emph{behavioural} analysis or \emph{structural} 
analysis.
The common practise of evaluating a model's accuracy 
on a targeted challenge dataset (designed to be representative of the feature or phenomenon 
in question) is 
a form of \emph{behavioural} analysis.
On the other hand, \emph{structural} analysis concerns the structural properties of the 
representation space, such as emergent feature clusters or geometric regularities.

A recently popular style of structural analysis is methodical diagnostic classification or 
\emph{probing\footnote{The term \emph{probing} is sometimes also used for studies featuring 
only behavioural analyses, but we refer specifically to the methodology surrounding 
auxiliary classification on top of intermediate model representations, as in \citet{alainbengio, hewitt-manning-2019-structural} and \citet{pimentel-etal-2020-information}}}. 
First introduced in \citet{alainbengio} as \emph{diagnostic classification}, probing
has been popular in the study of natural language processing models, where it has shown
the prevalence of rich syntactic features (for example in \citet{hewitt-manning-2019-structural}) and
semantic features \cite{vulic-etal-2020-probing}.
Given that the aim of probing studies is very different from the usual machine learning goal of 
achieving the maximum possible accuracy on test sets, there has been much discussion 
on the suggested best practices and possible pitfalls when carrying out probing studies 
\cite{zhangbowman, hewitt-liang, voita-titov-2020-information, pimentel-etal-2020-information, pimentel-etal-2020-pareto}. 
We attempt to take on most of the suggestions in our methodology, especially the control tasks and 
selectivity measure in \citet{hewitt-liang} and the complexity ranges in \citet{hewitt-manning-2019-structural, 
pareto}.

\section{Experimental Setup}

The experiments in this work are separated into two parts. The first part evaluates the accuracy of distinct models on the task of grounding UI elements for mobile devices, while the second part assesses the spatial reasoning capabilities of the models trained during the first stage. All the datasets generated, trained models and code used in this work are made available in \url{http://github.com/debymf/ipa_probing}.

The stages of the experiments are described in the following subsections.

\subsection{Interface Element Grounding}

\subsubsection{Datasets}
For this work, we focused on datasets containing commands for grounding mobile UI elements.
This choice was due to the more extensive availability of datasets, unlike desktop applications.
However, the concepts and experiments presented here can easily be transported to different interface environments, given their similar structure.
The datasets used for our experiments are PixelHelp and RicoSCA~\cite{li2020mapping}. 

RicoSCA is a dataset containing commands with its respective matching interface element, composed of textual content and bounding box.
The dataset contains a total of 329,411 instructions\footnote{The number of instructions is different from the reported value in the original paper since only the generation script is provided, and that does not ensure that the same dataset is generated. In order to ensure reproducibility, we will make available our version of the dataset with respective splits.} which we split between train/dev/set as shown in Table~\ref{tab:datasets}.
RicoSCA contains three types of commands, Name-Type, Absolute-Location and Relative-Location, which we map to extractive, spatial absolute and spatial relative reasoning, respectively.

PixelHelp contains 187 multi-step instructions, where 88 are general tasks, such as configuring accounts, 38 are e-mail tasks, 31 are browser tasks, and 30 photos related tasks.
PixelHelp is only used for testing, given its smaller size.
 In this work, we focus only on single-step commands; therefore, PixelHelp cannot be used in its original format. To obtain single-step commands from the multi-step ones found in PixelHelp, we used the \textit{action phrase extraction model} proposed by \citet{li2020mapping}. We obtained a total of 780 single step commands.
 This dataset does not contain annotations on the type of reasoning required for each command. To generate such data, two expert human annotators enriched the dataset with the correct types, obtaining a .78 Kappa score, providing strong reliability. The details of this dataset are also shown in Table~\ref{tab:datasets}. 
\begin{table}[!h]
\centering
\resizebox{\linewidth}{!}{%
\begin{tabular}{@{}lrrrr@{}}
\toprule
\multicolumn{1}{c}{\multirow{2}{*}{\textbf{Dataset}}} &
  \multicolumn{4}{c}{\textbf{Number of Instructions}} \\
\multicolumn{1}{c}{} &
  \multicolumn{1}{c}{\textbf{Extractive}} &
  \multicolumn{1}{c}{\textbf{Absolute}} &
  \multicolumn{1}{c}{\textbf{Relative}} &
  \multicolumn{1}{c}{\textbf{Total}} \\ \midrule
\textbf{RicoSCA - Train} & 100,921 & 6,531 & 57,222 & 164,674 \\
\textbf{RicoSCA - Dev}   & 40,504  & 2,529 & 22,577 & 65,610  \\
\textbf{RicoSCA - Test}  & 60,922  & 3,897 & 34,308 & 99,127  \\
\textbf{PixelHelp}       & 673     & 100   & 7      & 780     \\ \bottomrule
\end{tabular}%
}
\caption{Number of instructions for each dataset, indicating the type of reasoning required for grounding.  }
\label{tab:datasets}
\end{table}

\subsubsection{Models}

For the grounding of mobile UI elements, we compare the performance of four different models: BERT (base), RoBERTa (base), LayoutLM (base) and Seq2Act.

\emph{BERT}~\cite{devlin-etal-2019-bert} and \emph{RoBERTa}~\cite{liu2019roberta} are known to obtain excellent results in tasks where direct semantic similarity is required.
However, there is still no conclusive evidence that this model can reason spatially, especially for the task of grounding UI elements. 
\emph{LayoutLM}~\cite{xu2020layoutlm} is a ''Language" Model for the combination of text and document layout, geared towards document image understanding tasks.
This model takes as input a bounding box that denotes the relative position of a token within a document.
This positional embedding can capture the relationship among tokens within a document.
LayoutLM was initially proposed as a pre-trained model for physical documents, making this work a novel application for such a model.
This model can also take as input images representing a layout structure; however, we only consider the interface elements' positional features as input in this work.

In order to apply such pre-trained models, we reformulate the interface element grounding problem as a pairwise relevance classification problem. Given a natural language command $c$ and interface elements $U=\{u_1,\ldots,u_n\}$, a function $f(c,u_i)$ is defined, where $f(c,u_i)=1$ if $u_i$ is the target element in $c$ and $f(c,u_i) = 0$ otherwise. 

BERT and RoBERTa take as input the command $c$ and the text $u_{i_T}$ from the interface element being tested in the format $ [ c || u_{i_T} ] $. These models cannot receive the element's position as input since they cannot interpret such a structure.
On the other hand, LayoutLM also takes as input the bounding box $u_{i_{BB}}$ of the interface element, hence, it uses two different inputs: the textual information $ [ c || u_{i_T} ]$, in the same manner as BERT, and the positional information $[u_{i_{BB}}]$.

\begin{figure*}[!h]
    \centering
    \includegraphics[width=0.9\linewidth]{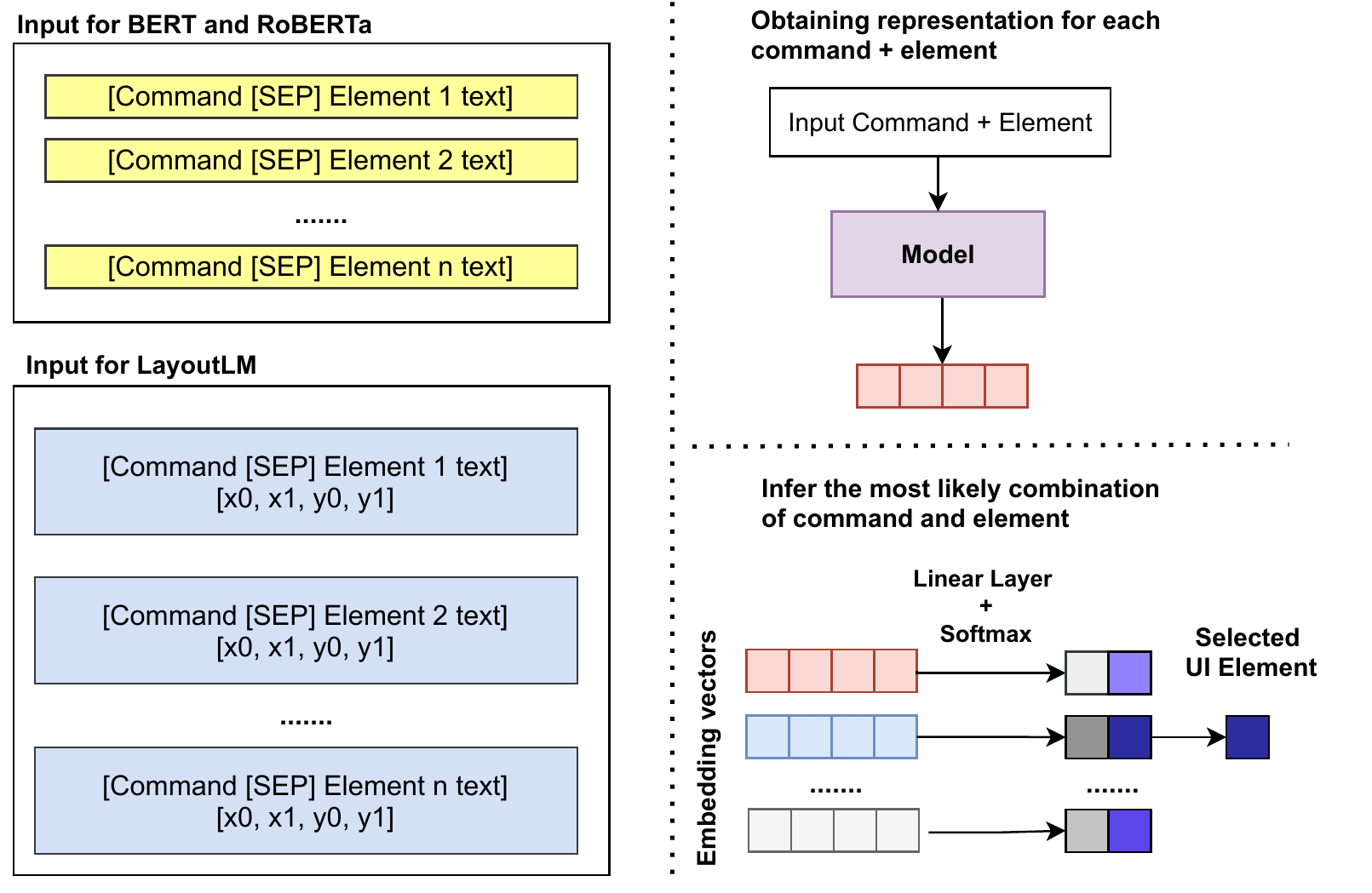}
    \caption{Inference pipeline for LayoutLM, BERT and RoBERTa. Both models generate representations based on the instruction and interface elements. The predicted interface element is inferred from the output probabilities for the pair classification setting. Best seen in colour.}
    \label{fig:inference_process}
\end{figure*}

These models are trained using the RicoSCA training and development set and tested for the PixelHelp and RicoSCA testing sets.
The supporting pipelines for inferring the correct interface element using these models are illustrated in Figure~\ref{fig:inference_process}.

Both models are trained\footnote{The models are trained with a learning rate of $1e$-$5$, batch size of $16$, a maximum sequence length of $256$ and $5$ epochs. Using $4$ Tesla V100s, both models take around $21$ hours to complete training.} with the objective to minimise Cross Entropy loss for the pair
classification task, so that the classifier is trained on the 
conditional likelihood $p(u_i \mid c)$ and select the UI element
$\argmax_{u_i \in U} ~p(u_i \mid c)$

The last compared model is \emph{Seq2Act}~\cite{li2020mapping}, composed of two different parts, an action phrase-extraction step and a grounding step. The first stage is a Transformer architecture which maps multi-step instructions to single-step ones. The grounding step matches these single-step commands to UI objects using another Transformer that contextually represents UI objects, and grounds object descriptions. Given the focus of this work, we are only interested in the results from the grounding stage. Seq2Act is trained using the full RicoSCA dataset; therefore, it was necessary to retrain the model using only the training split for RicoSCA. Given that the model was trained on a smaller split of the RicoSCA dataset, the obtained results differed slightly from the original paper.

\subsection{Probing Methodology}
Following recent works on the developing best practices in probing studies \cite{hewitt-liang, hewitt-manning, pimentel-etal-2020-information, pimentel-etal-2020-pareto, voita-etal-2019-analyzing}, we have followed three key
principles:
\begin{enumerate}
    \item Prioritising probe families with \emph{simple structure},
    \item Using a range of probing model complexities with in a family of probing models, and 
    \item Using randomised control tasks \cite{hewitt-liang} in order to ensure \emph{maximum selectivity} 
    for reported accuracies.
\end{enumerate}
\subsubsection{Probe Complexity Control}
For each auxilliary task, an MLP probe architectures, training 50 probes of varying complexities using the Probe-Ably framework of 
\cite{ferreira-etal-2021-representation}\footnote{\url{https://github.com/ai-systems/Probe-Ably} } with the default configurations for the MLP model. 
We use the number of parameters as a naive
approximation of model complexity. 

\subsection{Control Task and Selectivity Measure}
For the a given set of probe hyperparameters, we train the 
probes on a $30-20-50$ train-dev-test split of the auxiliary
task data.
For each fully trained probe, we report both the test accuracy
and the \emph{selectivity} measure \cite{hewitt-liang} in its
``entropic form": namely, the difference between the cross-entropy loss
of the auxiliary task test set and the cross-entropy loss
of the randomised \emph{control task} training set.
We use a simple control task where input representation vectors
have been assigned fixed random labels from among the auxiliary 
task target classes. 
Tracking the selectivity ensures that we are not using a probe that is
complex enough to be \emph{overly expressive} to the point of 
having the capacity to overfit the 
randomised control training set.

\subsubsection{Auxiliary tasks}

\begin{figure}[ht!]
\centering
\subfloat[Determining the absolute position.]
{ \includegraphics[width=0.40\linewidth]{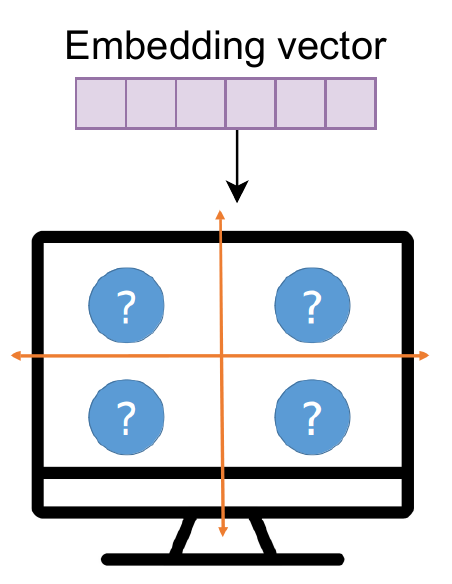} \label{fig:aux_task1}}
\qquad
\subfloat[Determining the relative position.]
{\includegraphics[width=0.40\linewidth]{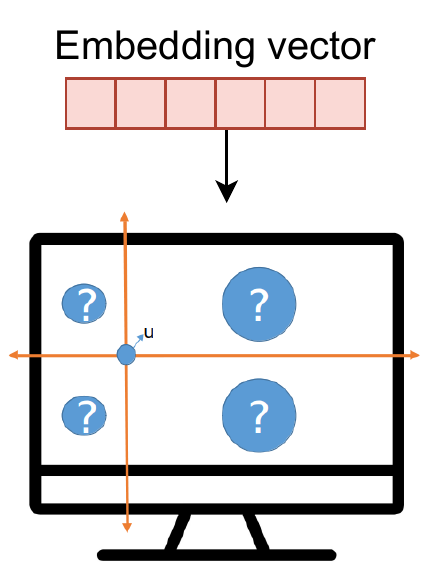} \label{fig:aux_task2}}
\caption{Illustration of the two types of auxiliary tasks, which requires determining the relative position from the obtained representation.}
\end{figure}

Towards probing the representation generated by LayoutLM, BERT and RoBERTa, we define four different auxiliary tasks, two for each spatial reasoning type:

\begin{itemize}
\item \textbf{Absolute spatial reasoning probing} (Figure~\ref{fig:aux_task1}): Given the representation of a command and an interface element, the first task is determining if the element is located at the bottom or top of the screen (\textbf{AT1}) while the second task focus on determining if the element is at the left or the right of the screen (\textbf{AT2}). 

\item \textbf{Relative spatial reasoning probing} (Figure~\ref{fig:aux_task2}): Given the representation of a command and an interface element, the third task requires positioning the interface element at the top or bottom of the element targeted by the command (\textbf{AT3}). Similarly, the last task indicates if the element is at the target's left or right (\textbf{AT4}).
\end{itemize}

\section{Results and Discussion}

This section presents the results for both parts of this work, the interface element grounding and the spatial reasoning probing.

\subsection{UI Grounding Performance}

Table~\ref{tab:results_grounding} presents the accuracy of all models for the task of interface element grounding. The results for the RicoSCA dataset are separated by the type of reasoning required: extractive reasoning (Ext), absolute spatial reasoning (Abs) and relative spatial reasoning (Rel).
We choose not to present the same for PixelHelp, given the small representativity of instructions demanding spatial reasoning in this dataset.

\begin{table*}[!ht]
\centering

\begin{tabular}{@{}llllllllll@{}}
\toprule
\multicolumn{1}{c}{\multirow{3}{*}{\textbf{Model}}} &
  \multicolumn{9}{c}{\textbf{Datasets}} \\
\multicolumn{1}{c}{} &
  \multicolumn{1}{c}{\multirow{2}{*}{\textbf{PixelHelp}}} &
  \multicolumn{4}{c}{\textbf{RicoSCA Dev}} &
  \multicolumn{4}{c}{\textbf{RicoSCA Test}} \\
\multicolumn{1}{c}{} &
  \multicolumn{1}{c}{} &
  \multicolumn{1}{c}{\textbf{Ext}} &
  \multicolumn{1}{c}{\textbf{Abs}} &
  \multicolumn{1}{c}{\textbf{Rel}} &
  \multicolumn{1}{c}{\textbf{All}} &
  \multicolumn{1}{c}{\textbf{Ext}} &
  \multicolumn{1}{c}{\textbf{Abs}} &
  \multicolumn{1}{c}{\textbf{Rel}} &
  \multicolumn{1}{c}{\textbf{All}} \\ \midrule
\textbf{BERT-Base} &
   86.28 &
   \textbf{99.61} &
   67.12 &
   54.75 &
   78.07 &
   \textbf{99.54} &
   67.07 &
   55.69 &
   77.92 
   \\
  \textbf{RoBERTa-Base} &
  \textbf{87.17} &
   99.60 &
   67.03 &
   56.31 &
   \textbf{78.12} &
   99.53 &
   68.03 &
   56.47&
   \textbf{77.98} 
   \\
\textbf{LayoutLM} &
  71.41 &
   98.51 &
   \textbf{86.75} &
   \textbf{61.40} &
   77.62 &
   98.40 &
   \textbf{90.80} &
   \textbf{60.53} &
   77.51 
   
   \\
  \textbf{Seq2Act } &
  75.86 &
  96.64 &
  67.14 &
  33.23 &
  73.95 &
  96.75 &
  65.89 &
  31.75 &
  73.78\\
 \bottomrule
\end{tabular}

\caption{Accuracy for the compared models. For the RicoSCA dataset, we compare the performance for different types of required reasoning for grounding. The Seq2Act results presented here are obtained from partial matching, since we are considering single-step commands.}
\label{tab:results_grounding}
\end{table*}

As expected, BERT and RoBERTa obtains outstanding performance for instructions requiring lexical overlap and semantic similarity. It easily outperforms the state-of-the-art model, Seq2Act, raising the importance of pre-trained model baselines in such studies: often, the basic similarity features captured by out-of-the-box language models are sufficient for complex tasks without needing to introduce new architectures.
Even though such models do not have access to the interface elements' position, they seems to manage (partially) the commands requiring spatial reasoning. 

Similarly, LayoutLM achieves almost perfect accuracy for extractive reasoning commands while also having a superior performance for absolute reasoning.
This performance is likely due to the fact that LayoutLM is integrated with BERT, with the added capability of representing positional information.

This result confers evidence that the structure learned from documents layout can also be transported to interface elements. LayoutLM achieves a lower performance on the relative spatial reasoning but still outperforms BERT for this task type. We hypothesise that such results could be improved by modifying the pair classification task to a selection across elements task. In this case, the model would be aware of all the screen elements when inferring a decision. However, such models are still very computationally demanding, and given that a screen can easily have up to $300$ UI elements, there is a need for more scalable approaches.

Even though Seq2Act is designed with the understanding of spatial reasoning in mind, we can notice that the model is outperformed in almost all aspects by transformers that encode only text. 

The overall obtained results highlight the need for more robust datasets for testing UI element grounding. Future work proposing new datasets and approaches capable of encoding positional information should initially ensure that the dataset evaluates such properties. 
The training protocol present in this work can be employed to remove trivial examples from the dataset. Instances of the dataset that can be predicted with high certainty by models that do not encode positions should be removed for testing specific spatial reasoning.

\subsection{Probing Results}

\begin{figure*}[!htbp]
\centering
\subfloat[AT1 probing.]
{ \includegraphics[width=0.90\textwidth]{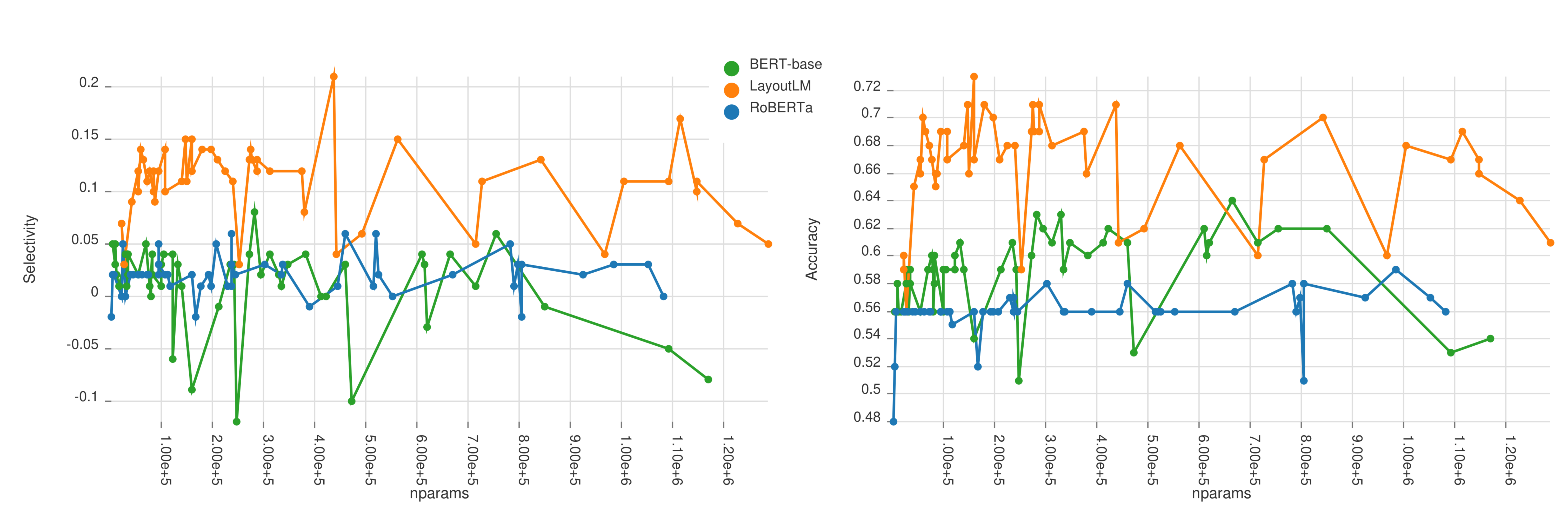}
	 \label{fig:plots_1_a}}
	 
	\subfloat[AT2 probing.]
{ 	\includegraphics[ width=0.90\textwidth]{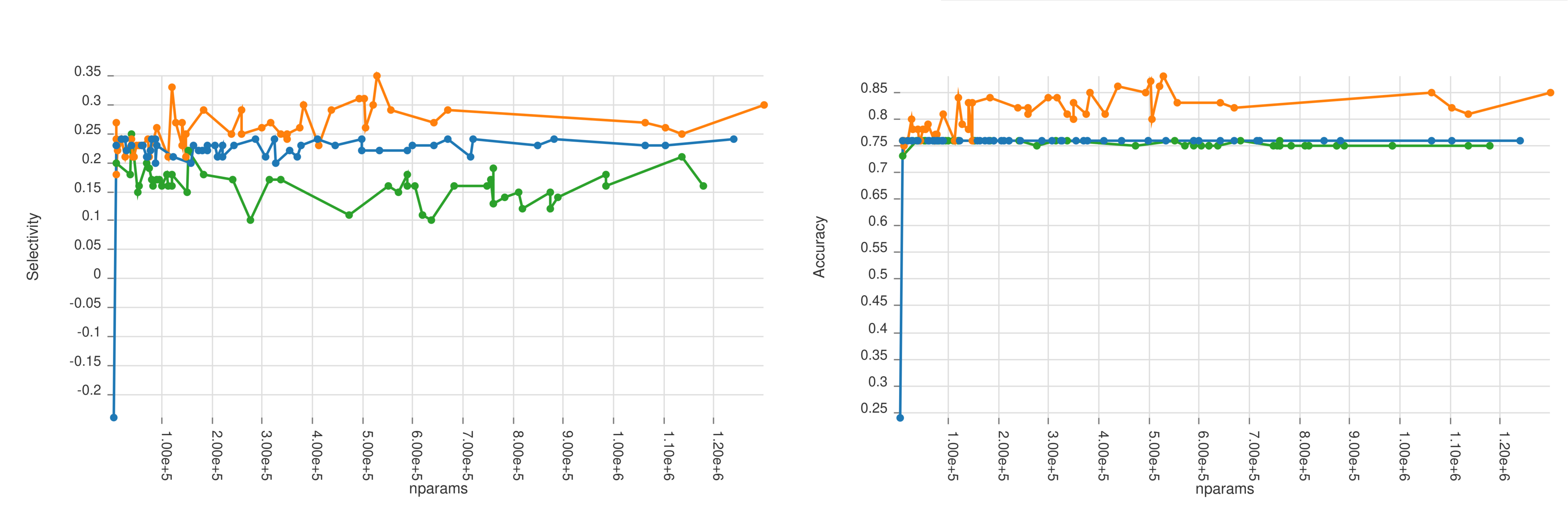}
	 \label{fig:plots_1_b}}
	 
\subfloat[AT3 probing.]
{\includegraphics[ width=0.90\textwidth]{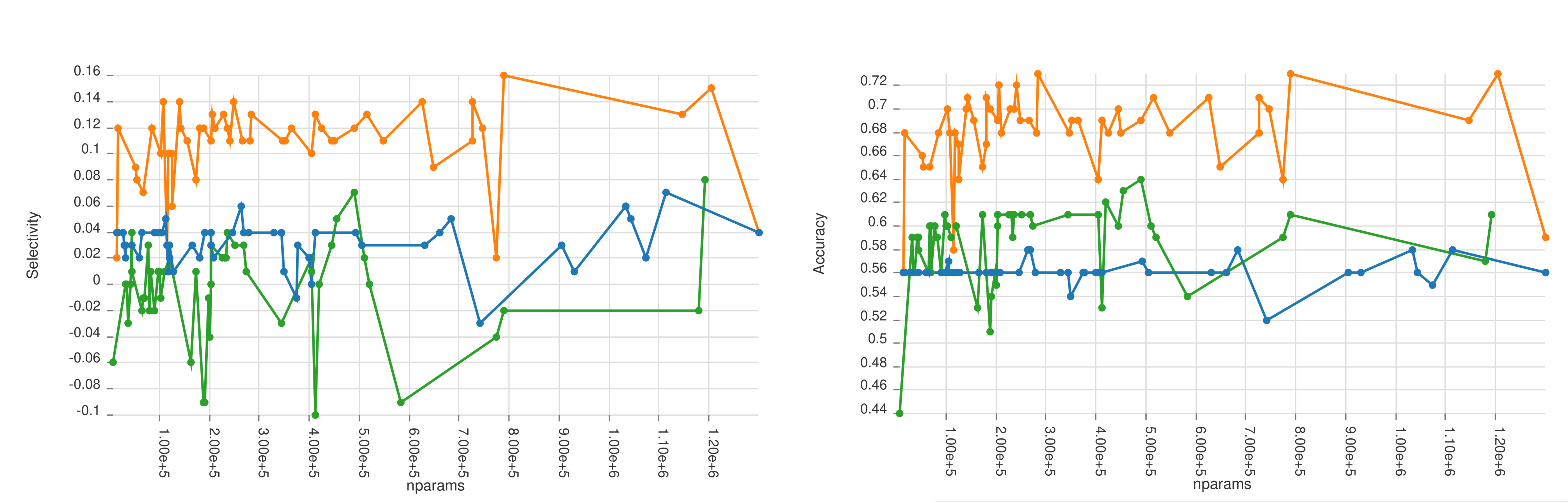}
	\label{fig:plots_2_a}}
		
\subfloat[AT4 probing.]
{
		\includegraphics[width=0.90\textwidth]{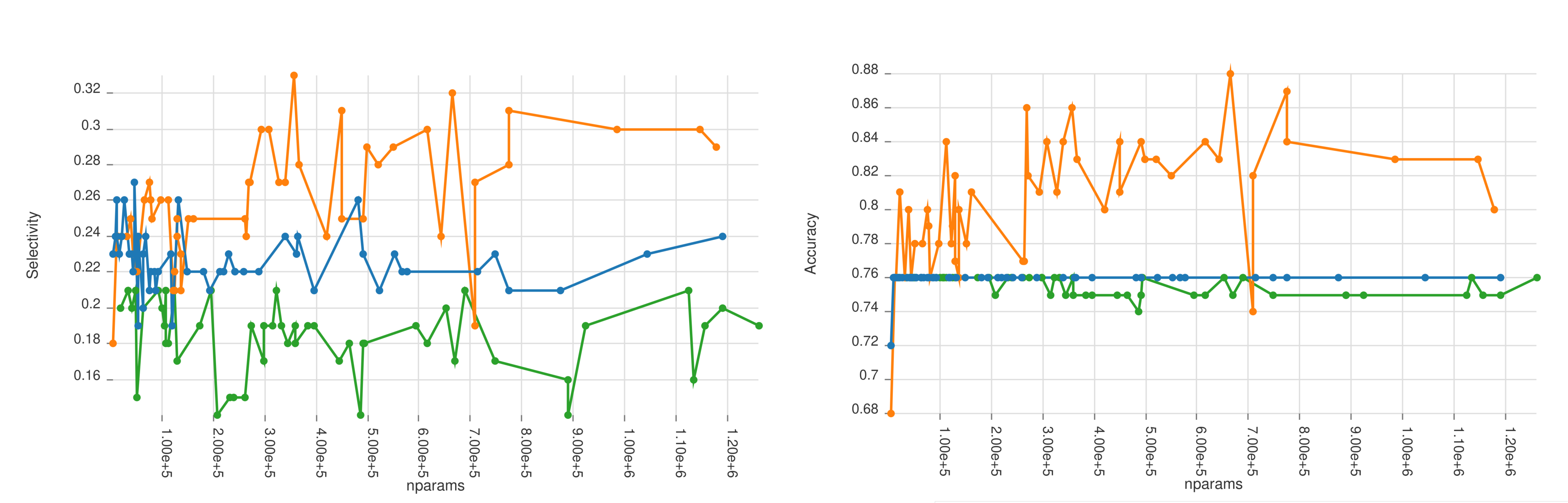} \label{fig:plots_2_b}}
\caption{Probing results for the both spatial reasoning task for MLP probes with varying complexity (approximated here by the number of parameters).}
\label{fig:all_plots}
\end{figure*}

In Figures \ref{fig:all_plots} (a-d) we present the
\emph{auxiliary task test accuracy} and the \emph{selectivity} of
50 trained MLP probes with varying complexity.
\footnote{Seq2Act does not generate a representation; hence, it was not included in the probing stage.}
Regions of maximum selectivity indicate that the accuracy scores are more trustworthy: low selectivity indicates little difference between the control task scores and the auxiliary task scores.
It is important to ensure that the selectivity doesn't drop off  (indicating an overly expressive probe), which is why the selectivity plots are included. 

As we can see from the MLP probes, LayoutLM outperform BERT and RoBERTa in the auxiliary spatial reasoning tasks, which shows that the generated representations are better able to discriminate between spatial labels.
This suggests its promising application for grounding UI elements and tasks with similar requirements, not limited to the domain of scanned documents. 

Overall, we can see that the Task  \textbf{AT2} was the easiest one for all models. In this task, the model was required to decide if an element is to the right or the left of the screen (considering a point in the middle). There is likely a bias in the dataset that makes it easier to distinguish between elements on the right or left of the screen. 

Again, such bias can influence the interpretation of the results, given the false interpretation that such models can understand positions such as left or right.  However, such a conclusion for BERT and RoBERTa is impracticable, given that they are unable to encode positions. The probing framework provided here can be applied to verify the quality of datasets used for grounding UI elements.

\section{Conclusion}

This paper presents an analysis of the UI grounding capabilities of three pre-trained models: LayoutLM,  BERT and RoBERTa. 
This analysis is performed in two parts: the first is a traditional analysis of the task accuracy, while in the second part, we perform a more profound analysis of the spatial reasoning performed by such models using probing techniques.

The results presented in this work highlights the need for datasets that go beyond lexical matching, focusing more on challenging spatial reasoning.
Even though RicoSCA contains a substantial amount of spatially focused commands, it is still a synthetic dataset with limited demonstration of the kinds of complex descriptions that would occur in natural language instructions.
There is a need for human-generated datasets of this kind, but such datasets are still challenging to obtain, especially those of a more natural variety 
(which is critical for future applications such as accessibility tools). 

Overall, We conclude that the greatest challenge in this area is the interpretation of instructions which require relative spatial reasoning that complicates the grounding task, binding it more tightly to a visual screen state.
While there is insufficient data to create strong models specifically for UI grounding of natural language instructions that include spatial descriptors, we have shown that document-based models such as LayoutLM learn a reasonable amount of spatially-relevant features (supported by the probing findings) that make them transferable to the UI grounding task. 

\bibliography{anthology, acl2021}

\end{document}